# Towards the Development of a Rule-based Drought Early Warning Expert Systems using Indigenous Knowledge

Adeyinka K. Akanbi
Department of Information Technology
Central University of Technology
Free State, South Africa
aakanbi@cut.ac.za

Muthoni Masinde
Department of Information Technology
Central University of Technology
Free State, South Africa
emasinde@cut.ac.za

*Abstract*—Drought forecasting and prediction is a complicated process due to the complexity and scalability of the environmental parameters involved. Hence, it required a high level of expertise to predict. In this paper, we describe the research and development of a rule-based drought early warning expert systems (RB-DEWES) for forecasting drought using local indigenous knowledge obtained from domain experts. The system generates inference by using *rule set* and provides drought advisory information with attributed certainty factor (CF) based on the user's input. The system is believed to be the first expert system for drought forecasting to use local indigenous knowledge on drought. The architecture and components such as knowledge base, JESS inference engine and model base of the system and their functions are presented.

## I. INTRODUCTION

The intricate complexity of drought has always been a stumbling block for drought forecasting and prediction systems [1]. This is mostly due to the web of environmental events (such as climate variability) that directly/indirectly triggers this environmental phenomenon. There are six broad categories of drought: meteorological, climatological, atmospheric, agricultural, hydrologic and water drought [1]. Nevertheless, irrespective of the category of drought, there is a consensus amongst scientist that drought is a disastrous condition of lack of moisture caused by a deficit in precipitation in a certain geographical region over some time period [2]. The effect of drought can be quantified based on the frequency, duration and intensity in the affected region subject to established timescales. For example, in South Africa, approximately 90% of the topography is either arid of semi-arid and drought is a characteristic feature of the climate [3]. Currently, the severity drought is so huge in the Western Cape region of South Africa with 2017 being the year with the lowest rainfall since 1933[1].

In recent years, scientists have developed different types of science-based knowledge to forecast drought conditions and predict likely behavior of climate using different types of drought indices and computational models [7,18]. However, data often used lacked the desired level of scalability, with other factors such as instrumental discrepancies and unlocalized dataset all contribute negatively to the degree of drought forecasting accuracy. This, in turn, affects the granularity of the projection information. On the other hand, local traditional farmers, use a different method to forecast drought. Preliminarily, before the advent of modern sciences, local farmers in different part of the world have developed their indigenous knowledge systems (IKS) which are based on the detailed observation of ecological and biodiversity interactions caused by climatic variation for their decision-making process. They use observation of natural indicators as guidance for drought prediction and forecasting. Once widely overlooked, the local indigenous knowledge is beginning to gain recognition [10]. Mostly because evidence suggests that no absolute error-prone methods for the drought forecasting and prediction that have been identified [2]. All current methodologies have an inherent subjectivity. For example, Palmer Drought Index [4], which is one of the most widely used index weakness lacks time-scale factor that does naturally exist with drought phenomenon [2].

Furthermore, experience has shown, however, that drought forecasting is a rather complicated process, hence, the use of IKS with modern science has been a subject of discourse [5-8]. Indigenous knowledge can be defined as the knowledge used by the local people in certain geographical area [9], and reflects unique expertise in the local environment. The indigenous knowledge on drought comprises of natural indicators/observation. In the KwaZulu Natal province of South Africa, the use of local indigenous knowledge on drought is still prominent for the region tribal farmers. The indicators include the timing of fruiting by certain local trees, blooming of certain florals, the water level in streams/ponds and the ecological behavior of animals. The indicators/observations are used by the domain experts to interpret weather/climate conditions to be expected. For example, to predict drought, IK expert holders formulate

---
[1] Statistical analysis of rainfall dataset from South African Weather Service (SAWS)

hypotheses about seasonal rainfall and other environmental conditions based on their observation of natural indicators. The deductive generation of drought knowledge implicitly through the natural indicators/observation may also be performed by an expert system using reasoning techniques.

The advancements in the field of Artificial Intelligence (AI) has provided the capability of developing information systems that have the abilities to generate inference, reasoning and prediction (from data which is provided). Information systems which have these abilities are called expert system. An expert system (ES) is a "*knowledge-intensive program that solves a problem by capturing the expertise of a human in limited domains of knowledge and experience*" [11]. Expert systems have been applied in a variety of areas, but not limited to, crop science, energy systems, economics (see section 5); and may fulfil a function that normally requires a human expert. Examples of the first generation ES are OPS5[2], KEE[3], and ART. For rule-based expert systems, domain expert knowledge is translated into a set of *rules* for reasoning and generating inference using appropriate reasoning strategies. In this paper, a Rule-Based Drought Early Warning Expert System (RB-DEWES) is presented. The uniqueness of RB-DEWES is its integrated knowledge base, which contains local indigenous expert knowledge on drought. The expert system would use probabilistic reasoning technique to automatically generate inference from the *rule set* for drought advisory information. The use of probabilistic reasoning would ensure a certainty factor[4] is attributed with the inferred information.

The rest of the paper is organized as follows: in Section 2 we introduce the background knowledge to rule-based expert systems. Section 3 describes the research methodology for the development of RB-DEWES. Section 4 deals with reasoning with uncertainty and in Section 5 we review some of the application of ES in other domains. In Section 6 we conclude this work and outline future work.

## II. RULE-BASED EXPERT SYSTEMS (RBES) AND THEIR APPLICATIONS

RBES uses human expert knowledge to solve real-life challenges in a specific domain. The domain-specific knowledge is stored in a knowledge base in form of *rules*; and are typically created by the knowledge acquisition engineer in conjunction with the domain expert. *Rules* are expert knowledge in the form of if-then conditional statements (modus ponens[5]). An inference engine component of the expert systems searches for a pattern in the input data that match patterns in the *rule set* to infer answers, generate predictions and make recommendations in the way a domain expert would. The "if" means when "the condition is true", the "then" means trigger a corresponding action. Hence, RBES require detailed information about the domain and the strategies for applying this information to problem-solving and generating inference.

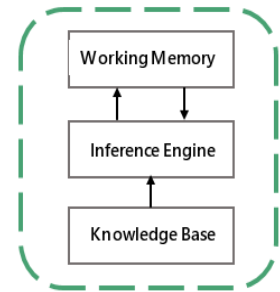

Fig. 1. Basic components of Rule-based System.

A typical RBES (Figure 1) comprises of three basic components. They are:
- the working memory,
- the inference engine, and
- the rule base.

### A. Rule base

Rule base (*knowledge base*) contains the set of *rules* which represent the knowledge of the domain [12]. The expert knowledge is represented in the form of "*if antecedents then consequent*". The *rule* base is used to generate inference from a sequence of pattern from the input data. The general form of a *rule* is:

---

IF Condition1 and
  Condition2 and
  Condition3
  ...
THEN Action1, Action2, Action3....

---

The conditions Condition(1-n) are known as antecedents. A *rule* is triggered if all antecedents (Condition(1-n)) are satisfied and consequents (Action(1-n)) are executed. However, some RBES allows the use of disjunctions such as 'OR' in the antecedents for complex scenarios before the Action(1-n) can be executed.

### B. Working memory (WM)

This is typically used to store the data input or information about the specific example of the problem or scenario. The WM is the database for storing collection of facts in a rule-based system and helps the system focus its problem solving [12].

### C. Inference engine

The function of the inference engine is deriving information or generating reasoning from a given problem using the *rules* in the knowledge base. The inference engine must find the right facts, interpretations, and *rules* and assemble them correctly. The two basic methods for processing the *rules* are –

---

[2] See https://en.wikipedia.org/wiki/OPS5
[3] See https://en.wikipedia.org/wiki/Knowledge_Engineering_Environment
[4] Measure of belief and disbelief.
[5] The rule of logic which states that if a conditional statement ('if p then q') is accepted, and the antecedent (p) holds, then the consequent (q) may be inferred.

Forward Chaining (data-driven, antecedent-driven) and Backward Chaining [12]. For example, in forward chaining all the facts are input to the systems and the system makes a deductive inference based on the *rules* available in the *rule set*. A system exhibits backward chaining if it tries to support a hypothesis by checking the facts in the *rule base* trying to prove that clauses are true in a systematic manner.

## III. DEVELOPMENT OF RULE-BASED DROUGHT EARLY WARNING EXPERT SYSTEM (RB-DEWES)

A Rule-based expert system is an expert system that uses to derive conclusion(s) from user inputs [22]. The system consists of a database that stores the domain acquired facts, the *rules* for inferring new facts and the inference engine (rule interpreter) for controlling the inference process. A *rule* consists of a premise and action. Since most domain experts formulate their knowledge in the form of *rules*. This makes *rules* the most widely used knowledge representation in expert systems. Therefore, the goal of the expert system is reproduction of the knowledge and reasoning capabilities of the domain experts by formalizing their knowledge for implicit reasoning. The is achievable by ensuring relevant knowledge are explicitly formulated, while any derivation from the knowledge is subjected to explicit *rules*. The purpose of this research was to design and develop a rule-based drought early warning expert system for predicting/forecasting drought using local indigenous knowledge. This research is part of an ongoing project towards the semantic integration of heterogeneous knowledge sources in the environmental monitoring domain [5].

In this section, the development methodology of RB-DEWES is presented. The overview of the development framework is shown in Figure 2. There are 4 (*four*) phases of the expert system development methodology. Phase 1 is the knowledge engineering phases that comprises of the knowledge acquisition, knowledge categorization and knowledge representation. Phase 2 deals with the architecture of the system. Phases 3 is the system design and development phase. In phase 4 we describe the system operation and evaluation. The detailed description of these phases is described as follows:

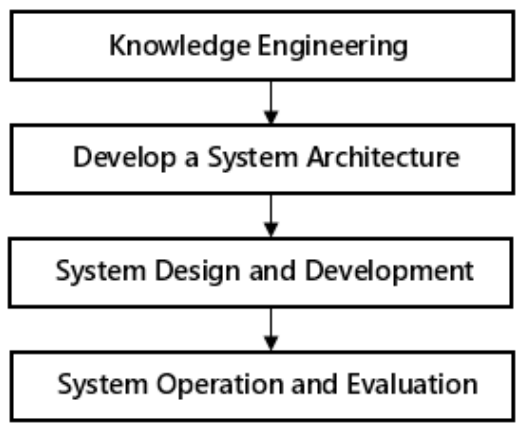

Fig. 2. RB-DEWES System Development Methodology.

### A. Knowledge engineering

This phase deals with the knowledge engineering part of the system development. It consists of the knowledge acquisition from the domain expert, categorization of the knowledge and the representation of the knowledge in form of *rules*. Each process is detailed as follows and depicted in Figure 3:

*1) Knowledge acquisition:* The process of knowledge acquisition (KA) is the most difficult phase of expert system development and its also time consuming [14]. However, it is the first step in the development of any knowledge-based systems. KA is the process of acquiring and transferring knowledge from the knowledge source (*domain expert*) to knowledge engineer (*expert system builder*) [13]. Through years of experience, tribal farmers have developed unique expertise in the local environment. They have been able to achieve this by studying the ecological interaction in the environment with respect to climate changes. As a result of the knowledge acquired, the domain experts could pinpoint the onset of drought from environmental observation. Many techniques have been developed for KA. However, irrespective of the technique, the first step is 'domain acquaintance'. This step allows the knowledge engineer to understand the domain with regards to the terminologies and get acquitted on formulating the key problem of the domain. This can be achieved by studying related literature and active learning. The next step is the KA, which can be in the form of interviews with domain experts, the use of a questionnaire to obtain implicit knowledge and focus group meetings. The process of KA from the domain expert could either be through – direct KA or indirect KA. The prevailing approach to KA in this research is through direct KA [12]. This is due to the fact that indirect KA is prone to error and expensive [12]. The direct KA process consists of an expert questionnaire, interviews, focus group meeting and the use of mobile application on smart devices.

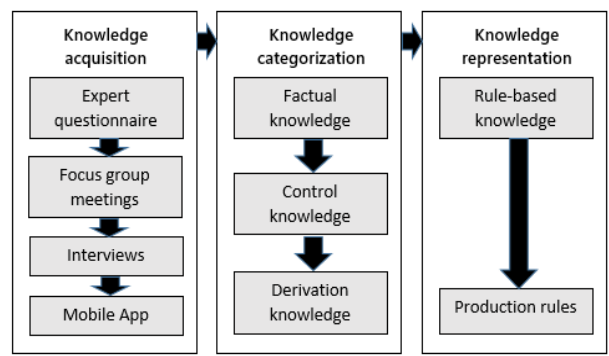

Fig. 3. The process flowchart of knowledge engineering phase.

*a) Surveying and Interviewing a human expert:* The use of questionnaires and interviews were aimed to understand and acquire the local indigenous knowledge on drought, the natural indicators, ecological observations and criteria for determining onset of drought. In this research, 21 domain experts from Swayimane, KwaZulu Natal province took part in the KA process. The expert questionnaire included 32 questions related to metrological, astronomical, weather and climate knowledge on drought. Also, structured and unstructured interviews were used. The questionnaire consists of the following components:

- The first component aimed to acquire the respondent's knowledge on weather forecasting.
- The second component aimed to document the effectiveness and use of indigenous weather forecast cropping decisions.
- The third component aimed to identify and document the unstructured weather indicators based on the categories and patterns of the seasons, ecological interaction, astronomical, meteorological and animal/plants behavior with application examples. This component provides an insight into the formulation of domain *rules* based on years of experiences from observation.

*b) Focus group meeting and mobile application:* IK was also obtained from the focus groups comprising of 6-10 domain experts in a series of oral consultation, meeting sessions through a local intermediary. The unstructured IK are manually captured. Also, an Android application was specifically developed for the purpose of KA in this research. The mobile app pictorially captures the natural indicator/observation, the description and spatial coordinates. These data remotely stored in the IK Web App Server or IK database server (Figure 4).

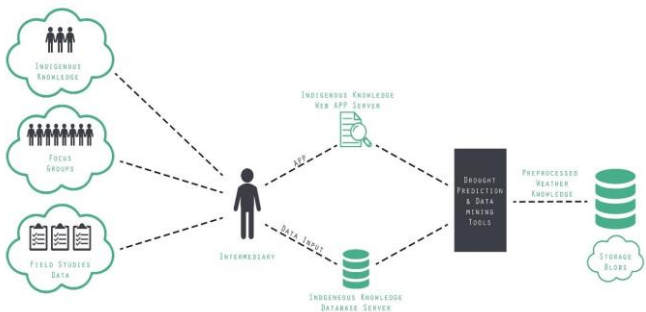

Fig. 4. The Indigenous knowledge acquisition framework.

*2) Knowledge Categorization:* Based on the origin of knowledge, distinctions were made in categorizing domain expert knowledge according to its use. The knowledge relevant to drought forecasting and prediction was categorized into three types: factual knowledge, control knowledge and derivation knowledge [14]. The rules for the RB-DEWES are derived from the derivation knowledge. The derivation knowledge was distinguished in physical observation and was detected by a domain expert. For example, the sighting of Ntuthwane ant and high soil moisture indicates the recent occurrence of rainfall. Control knowledge is the meta-rules that control the deductive rules from the derivation knowledge. The factual knowledge was distinguished by experience and decision making. Such as the experience gained from the application of rules (derivation knowledge) to make decisions.

*3) Knowledge Representation:* The aim of this phase in the system development is to encode the expert knowledge on drought. This is achieved by the formalization of the facts and relationships that constitute the expert knowledge on drought into a machine-readable format. Although there exists different methodology for knowledge representation such as ontology, semantic networks, logic expressions, casual networks and *rules* [15]. However, rule-based programming style has become very popular for expert system due to astonishing deductive inference performance by combining relatively few *rules*. The rule base for RB-DEWES contains about 32 natural indicators (astronomical observations, meteorological observations, animal behaviors and plant behaviors).

TABLE I. INDIGENOUS ANIMAL, PLANTS, METEOROLOGICAL, ASTRONOMICAL INDICATORS INCLUDED IN THE EXPERT SYSTEM

| Animals | Plants | Meteorological | Astronomical |
|---|---|---|---|
| Magwababa bird | Mviti tree | Humidity | Full Moon |
| Inkonjane ant | Wiki-Jolo tree | Soil moisture | Half Moon |
| Ntuthwana ant | Umphenejane tree | Weather temperature | Stars |
| Ingxangxa frog | Peach tree | Rainfall | Day Sky |
| Onogolantethe bird | Amapetjies tree | Thunderstorm | Night Sky |
| Phezukomkhono bird | Tshi tree | Sunlight Intensity | |
| Cows | Motoma tree | Windstorm | |
| Inyosibees | Marakarakane tree | | |
| Lehota frog | Mutiga tree | | |
| All_animals | All_plants | | |

*B. Architecture and components of RB-DEWES*

The system architecture provides an overview of the system development process and serves as a guideline for outlining the functionalities of the system. We have developed the following architecture for RB-DEWES (Figure 5), which includes five main components: (1) a graphical user interface, (2) a database, (3) an inference engine, (4) a knowledge base and (5) model base. Figure 5 below shows the basic architecture of RB-DEWES.

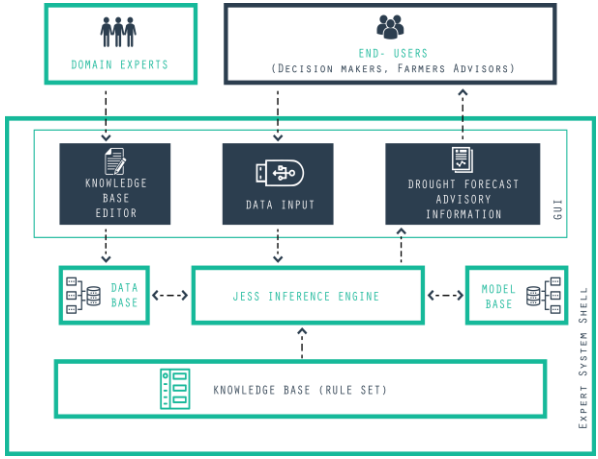

Fig. 5. The architecture of RB-DEWES.

*1) Graphical User Interface (GUI):* The interactive interface allows the communication between the domain expert, end-users (decisions makers, farmer's advisors) and the RB-DEWES. The system is designed to be users friendly for interaction with domain expert who are saddled with creating or editing the KB; and with the end-users of the ES. Different interfaces are designed for knowledge base editor, data input with CF and Drought forecast Advisory (DFA) output with CF.

*2) Database:* The database component of the system utilizes a SQL server which is a relational database to store the indigenous knowledge on drought. The database is used to store the facts, data required for the knowledge base and drought mitigation information that might be applicable after the deductive inference. Domain-based knowledge was collected with questionnaires, focus droup meetings, mobile app and were translated into SQL data by system developers.

*3) Inference engine:* The inference engine performs rule-based reasoning using forward chaining technique. This component contains the software code which processes the *rules* based on the facts given by domain expert for a given situation. It predicts the likely onset of droughts based on the rule-patterns experience stored the knowledge base and issue a drought advisory knowledge information via the GUI. RB-DEWES makes use of Java Expert System Shell (*JESS*).

*4) Knowledge base:* The knowledge base is used to store the domain knowledge represented in form of *rules* required for drought forecasting and prediction. The local indigenous knowledge on drought comprises of observation of different environmental phenomenon and natural indicators of ecological interaction are represented as object-attribute-value (*O-A-V*) and shown in Table 2. The following sample *rules* are for determining the likely onset of drought using local indigenous knowledge from KwaZulu Natal province:

RC18: *IF rainfall is High*
 *AND soil moisture is high*
 *AND soil temperature is moderate*
 *THEN no evidence of drought (0.9)*

RC21: *IF phezukomkhono is sighted*
 *AND Guavatree is flowering*
 *AND Wiki-Jolo is blooming*
 *AND Umphenjane is flowering*
 *THEN No evidence of drought, onset of spring (0.85)*

RC30: *IF mviti tree is flowering*
 *AND weather temperature is high*
 *AND ntuthwane ant was sighted*
 *AND soil moisture is low*
 *AND amapetjies is flowering*
 *THEN No evidence of drought, onset of summer (0.70)*

RC15: *IF mviti shows wilting*
 *AND Inyosibees is sighted*
 *AND Moon appears full*
 *THEN moderate evidence of drought, onset of autumn (0.75)*

RC2: *IF umphenjane is blooming*
 *THEN no evidence of drought (0.4)*

RC5: *IF soil moisture is high*
 *THEN no evidence of drought (0.5)*

RC6: *IF phezukomkhono is sighted*
 *THEN no evidence of drought (0.6)*

RC10: *IF humidity is high*
 *THEN no evidence of drought (0.6)*

RC38: *IF all_animals are thin*
 *AND all_plants shows wilting*
 *AND humidity is high*
 *AND rainfall is none*
 *AND day sky appears clear*
 *AND night sky is clear*
 *AND stars are sighted*
 *AND weather temperature is high*
 *AND sunlight intensity is high*
 *THEN evidence of drought (0.68)*

TABLE II.  INDIGENOUS KNOWLEDGEON DROUGHT NATURAL INDICATORS AND OBSERVATION IN O-A-V FORM.

| Rule Condition | Object | Attribute | Value | CF |
|---|---|---|---|---|
| RC10 | Umphenjane | Is | Blooming | 0.40 |
| RC5 | Soil moisture | Is | High | 0.50 |
| RC6 | Phezukomhkono | Is | Sighted | 0.60 |
| RC10 | Humidity | Is | High | 0.60 |
| RC15 | Mviti | Shows | Wilting | 0.75 |
| RC15 | Inyosibees | Is | Sighted | 0.75 |
| RC15 | Moon | Appears | Full | 0.75 |
| RC17 | All_animals | Appears | Thin | 0.50 |
| RC17 | All_plants | Appears | Wilting | 0.50 |

*5) Model base:* This is the component of the expert system that executes the MYCIN model for modelling uncertainty in the system. A qualitative probabilistic model attribute certainty factor to the inferred output. This is the measure of the expert system confidence level based on the user's input data or set of input data.

*C. RB-DEWES System Design and Development*

This application is designed to be used by decision-makers, farmer's advisors, farmers call centers in providing advisory information on drought to farmers. The inference engine will be programmed using JESS, and SQL server 2012 was used as the database platform. RB-DEWES runs on Microsoft Windows platform and is made compatible with Microsoft Windows Server OS or Microsoft Windows PC OS. The relational database contains the detailed indigenous knowledge on drought and drought mitigation strategies. While the knowledge base consists of *rules* for solving domain-specific problems. An exhaustive review of the available software for developing expert system shell was conducted prior to the final selection of JESS. JESS is a rule-based tool for building intelligent expert systems. It was developed in 1995 by Ernest Friedman-Hill of Sandia National Labs, USA [16]. JESS engine is implemented in Java, and can be referenced by Java object.

The hardware and software minimum requirement of RB-DEWES are as follows:

*1) Hardware:* The minimum hardware component to run the system are:
- A PC with Intel CPU, 4GB RAM and 1GB hard disk space.
- A VGA monitor

*2) Software:* The minimum requirement for the software component for the runtime environment and to be installed are:
- Microsoft Windows 10
- SQL Server 2012
- JAVA SE Runtime Environment 7

*D. Expert system operation and evaluation*

Running on the Windows PC/Server, at the start of each forecasting(predicting) session, the user is prompted to select a role from the GUI to determine the appropriate interface from the three available interfaces – knowledge base editor, data input and output. The user operates the system through the GUI and supplies data using push buttons, radio buttons, drop-down list, text –field etc. The knowledge base editor interface allows the domain expert to add, edit, and delete *rules* and other contents in the KB and database.

The data input interface allows pre-defined observation and natural indicators with the certainty factor to supplied as input by the end-users based on his or her personal observation for deductive inference. Multiple observation or occurrence(s) of natural indicators can be selected from the list. The inference engine generates the inference based on the *rules* in the knowledge base. After each inference, the DFA information is generated as output with attributed CF; indicating the system level of certainty based on users input and *rules*.

After the deductive inference from the users' input, the inferred information is displayed through the output interface. Additionally, drought mitigation information stored in the database is displayed to the end-user depending on the severity of the drought. This added information could help end-users in adopting the suitable mitigation strategy.

Expert system evaluation is the process of determining the quality of inferences generated by the expert system [13]. According to [18,19], the process consists of verification and validation. Verification ensures the knowledge captured in the system accurate and correctly implement its specifications; while validation process involves verifying that a system performs at the satisfactory level of accuracy according to system requirement. However, both process will form the basis of our future work.

IV. REASONING WITH UNCERTAINTY

Determining the level of certainty in decision-making programs is very critical. In an expert system, the vagueness of expert *rules* and ambiguities in users input are the major factors affecting the absolute certainty of system outputs. Hence, an expert system must exhibit a high level of modularity, and each *rule* may have associated with it a certainty factor (CF). The CF is a measure of the confidence in the piece of knowledge. However, there are many ways in which CFs can be defined and combined with the inference process. Our system incorporates the MYCIN model [17] for calculating the CF. The model ensures the *rule* probability is calculated by multiplying the domain expert implication probability by the user's input precondition probability. The domain expert implied probability is stated in the *rule* and expresses the expert confidence level based on a set of condition(s). On the other hand, the user's input precondition probability determined by the user is also utilized. The CF value was calculated applying (1).

$$P = P_{old} + (1 - P_{old}) * P_{new} \quad (1)$$

For example, considering a *rule* (R25) for generating inference based on the combination of set of observation and natural indicators (preconditions):

TABLE III. RULE SAMPLE

| Rule No | IF | Relation | THEN | CF |
|---|---|---|---|---|
| R25 | RC2 | && | | |
| | RC5 | && | | |
| | RC6 | && | | |
| | RC10 | && | No evidence of drought | 0.80 |

Suppose, the user input data (UID) the following preconditions and their corresponding CF values through the system GUI.

TABLE IV. USER INPUT DATA SAMPLE

| UID | Object | Attribute | Value | Relation | CF |
|---|---|---|---|---|---|
| UID2 | Umphenjane | Is | Blooming | && | 0.90 |
| UID5 | Soil moisture | Is | High | && | 0.50 |
| UID6 | Phezukomkhono | Is | Sighted | && | 0.80 |
| UID10 | Humidty | Is | High | && | 0.70 |

In this example, the relation of the preconditions is "AND". The probability of the preconditions is given by the minimum of the precondition set *CF* i.e. min[UID2(CF), UID5(CF), UID6(CF), UID10(CF)]. Else, it will be maximum if the relation is "OR". Therefore, based on the given example, probability of the preconditions is: min(0.9,0.5,0.8,0.7) = 0.5. The CF of the inferred knowledge based on the R25 will be as 0.8*0.5=0.4=40%.

Therefore, the model base will attribute a CF value of 40% to the inferred knowledge displayed as output. However, if the old CF value for "No evidence of drought" derived from another *rule* is 90%. The new CF is: 0.9 + (1 - 0.9) * 0.4 = 0.94 = 94%.

## V. RELATED WORK

The literature is awash with the use of rule-based expert system in numerous application areas. For example, applications of rule-based systems are including financial fraud prediction system, chemical incident management, medical treatment, power transmission protection etc. However, the exists no implementation of ES in drought forecasting using local indigenous knowledge.

[13] proposed a diagnostic advisory rule-based expert system for integrated pest management in solanaceous crop systems called DIARES-IPM. They aim the system will help non-experts to identify pests and suggest appropriate treatments. Their system expert system shell was EXSYS for Windows. The rules in the EXSYS are made up of "Quantifiers", "Choices", and "Variables". In addition to the rule-based knowledge representation, knowledge is also represented in form of pictorial illustration.

In a similar vein, [20] developed a new methodology for providing intelligence in the system for diagnosis of diseases of oilseed-crops, such as soybeans, groundnut and rapeseed-mustard. The system called WIDDS is based on fuzzy logic reasoning technique. They believe the system would generate inference faster than traditional approach and will serve as an effective decision-making tool. The web-based system generates inference from diseases knowledge for the aforementioned oilseed crops.

In the domain of fish disease management, [21] developed a fish-expert which is a web-based expert system for fish diseases diagnosis. The system has over 200 rules and hundreds of images for different types of fish diseases. The system has been tested and currently in use by fish farmers in the North China region.

## VI. CONCLUSION AND FUTURE WORK

The RB-DEWES provides an effective tool to improve drought forecasting and prediction using local indigenous knowledge. The system employs rule-based methodology and probabilistic reasoning technique. Our research was motivated by the need to utilize local indigenous knowledge in conjunction with modern AI techniques. Therefore, the designing and development of a rule-based drought early warning expert systems have been presented in this paper as a platform towards the implementation of an expert system for drought forecasting using local indigenous knowledge. This approach facilitated the generation of inference with enhanced intelligence from knowledge acquired from domain experts. The system allows the attribution of certainty factors to the input and output information, which tremendously helps with evaluating the quality and confidence level of the system. It is able to mimic the indigenous domain expert drought forecasting processes by matching users input from observing ecological interactions and natural indicators with rule set patterns.

However, there are some inherent limitations of the system, such as:

- Currently, the *rules* for RB-DEWES are based on the indigenous knowledge acquired from domain experts in Swayimane, KwaZulu Natal province, South Africa. Further knowledge acquisition could be carried out in other geographical areas to diversify and expand the knowledge base to achieve a comprehensive rule-based drought early warning expert system.

- In increasing the reliability of drought forecasting systems. It is not enough to consider local indigenous knowledge domain alone; effort must be made in the future towards the integration of inferences generated from the indigenous knowledge domain with the modern drought forecasting model information.

Hence, our future work are as follows:

- Improving the mechanism of drought early warning system through the application of an ontological-based reasoning technique.

- The semantic representation and integration of inferences generated from heterogeneous knowledge bases for a more accurate drought forecasting and prediction system.

- In its present version, RB-DEWES is a standalone, designed to operate on a PC/Server environment. Future implementation effort will make it available as a web-based expert system with qualitative system evaluation.

ACKNOWLEDGMENT

We would like to thank many indigenous knowledge domain experts from Swayimane, KZN province, South Africa,

that participated in this research for their cooperation and support. Special thanks should go to Dr. Tafadzwanashe Mabhaudhi for facilitating the knowledge acquisition component of this research.